\def\year{2021}\relax
\documentclass[letterpaper]{article} 
\usepackage{aaai20}  
\usepackage{times}  
\usepackage{helvet} 
\usepackage{courier}  
\usepackage[hyphens]{url}  
\usepackage{graphicx} 
\usepackage{array}
\usepackage{tabularx}
\usepackage{booktabs}
\usepackage[table]{xcolor}
\usepackage{hyperref}

\usepackage{tikz}       
\usepackage{subfig}
\usepackage{multirow}   
\usepackage{bchart}     
\usepackage{hyperref}
\usepackage[T1]{fontenc} 

\definecolor{Gray}{gray}{0.9}

\definecolor{darkblue}{rgb}{0, 0, 0.5}
\hypersetup{colorlinks=true,citecolor=darkblue, linkcolor=darkblue, urlcolor=darkblue}

\usepackage{pifont}
\newcommand{\cmark}{\ding{51}}%
\usepackage[show]{notes}
\newcommand{\ignore}[1]{}
\newcommand{\com}[1]{}
\newcommand{\oren}[1]{\inote[oren]{{\color{red} #1} \color{black}}}

\newcommand{\rev}[1]{\textcolor{black}{\rm #1}}

\usepackage[super]{nth}

\title{Discourse Parsing of Contentious, Non-Convergent Online Discussions}

\author{Stepan Zakharov,\\
{Ben}\\
email@wolfsburg.com,
\And
Alfonso Alpaca,\\
{Alexandria}\\
email@alexandria.com, 
\And
Carl Capybara,\\
{Canberra}\\
email@canberra.com}

\author{Stepan Zakharov,\textsuperscript{1,\dag} Omri Hadar,\textsuperscript{2,\ddag} Tovit Hakak,\textsuperscript{3,\ddag} Dina Grossman,\textsuperscript{4,\ddag} \\
\AND 
Yifat Ben-David Kolikant,\textsuperscript{5,\ddag} Oren Tsur\textsuperscript{6,\textdagger}\\
  \textsuperscript{\dag}Software and Information Systems Engineering\\ Ben Gurion University  \\
   \textsuperscript{\ddag}School of Education \\The Hebrew University of Jerusalem\\
  \texttt{\textsuperscript{1}stepanz@post.bgu.ac.il}\\
  \texttt{\{\textsuperscript{2}omri.hadar,\textsuperscript{3}tovit.hakak,\textsuperscript{4}dina.grossman,\textsuperscript{5}yifat.kolikant\}@mail.huji.ac.il}\\ \texttt{\textsuperscript{6}orentsur@bgu.ac.il}
}

\begin{document}
\maketitle
\begin{abstract}
Online discourse is often perceived as polarized and unproductive. While some conversational discourse parsing frameworks are available, they do not naturally lend themselves to the analysis of contentious and polarizing discussions. Inspired by the Bakhtinian theory of Dialogism, we propose a novel theoretical and computational framework, better suited for non-convergent discussions. We redefine the measure of a successful discussion, and develop a novel discourse annotation schema which reflects a hierarchy of discursive strategies. 
We consider an array of classification models -- from Logistic Regression to BERT. We also consider various feature types and representations, e.g., LIWC categories, standard embeddings, conversational sequences, and non-conversational discourse markers learnt separately. Given the 31 labels in the tagset, an average F-Score of 0.61 is achieved if we allow a different model for each tag, and 0.526 with a single model. The promising results achieved in annotating discussions according to the proposed schema paves the way for a number of downstream tasks and applications such as early detection of discussion trajectories, active moderation of open discussions, and teacher-assistive bots. Finally, we share the first labeled dataset of contentious non-convergent online discussions.  
\end{abstract}

\section{Introduction}
\label{sec:intro}
Online discussions are often perceived as polarized and unproductive. Empirical evidence, gained from field experiments in collaborative learning, suggests that raising the awareness of the participants to the dynamics of the discussion influences their discursive behavior, ultimately making the discussion more productive \cite{wise2011analyzing,yoon2011using,chen2018fostering}. However, reflecting on conversational dynamics in real time is challenging and often requires experienced human moderators. In order to mitigate this challenge, we had developed a novel annotation scheme for discourse parsing. Our annotation scheme, inspired by the Bakhtinian principles of \emph{dialogism}, is explicitly designed to meets two objectives: (i) capturing strategic discursive moves that impact the trajectory of a discussion, and (ii) acknowledging that a successful and productive discussion does not necessarily mean that participants converge to agreement. 

Existing schemes, for the most part, rely on {\em convergence} as a measure for the success (productiveness) of a discussion. Success is declared if a convergence to consensus is achieved, e.g. \cite{chiu2000group,diziol2010using,schwarz2018orchestrating}. Consequently, computational work addressing conversational discourse typically aim at modeling discussions in which convergence/agreement is the declared goal. Two examples of this setting are customer service log calls \cite{jurafsky1997switchboard}, and problem solving of a math assignment in a collaborative learning environment \cite{schwarz2018orchestrating}.
 
Many discussions, however, are often polarized, contentious, and {\em non-convergent} in nature. We argue that existing schemes for discourse analysis do not properly capture the dynamics exhibited in fruitful, yet non-convergent discussions, and therefore cannot be used effectively to evaluate or improve the quality of online discourse. 

In order to address this problem, we redefine the success of a discussion in a way that accounts for polarized, non-convergent discussions. We further introduce a new annotation scheme for non-convergent discussions, and define a novel discourse parsing task. Our annotation scheme is inspired by the Bakhtinian concepts of Internally Persuasive Discourse (IPD) and Dialogic Agency -- a crucial capacity in a modern, liberal, multi-cultural society. Another key factor guiding the development of the annotation schema is the observation that conversations unfold sequentially, thus a speaker's utterance is triggered only by previous utterances.  While this observation may appear trivial, most previous work on discourse parsing analyze conversations as a whole.
In this work we take a different approach. First, we focus on the discursive evolution of an ongoing conversation, rather than on an isolated utterance or on the complete conversation. Accordingly, our proposed tagset and the annotation process, were tailored to capture conversational strategies used by speakers (and perceived by participants and annotators) at the time of the utterance, and not in hindsight, after the conversation is complete. We elaborate on the theoretical background and the novel parsing tasks in Sections \ref{sec:theory} and \ref{sec:annotation_scheme}.

In line with the motivation stated above, the proposed annotation scheme poses two main computational challenges: (i) it aims at discourse parsing at {\em real time}, as the discussion evolves, thus the available context is limited, and (ii) the tagset contains 31 tags and each utterance can (sometimes should) be tagged with a set of tags, rather than a single tag, making sequence optimization hard. We address the computational challenges in Section \ref{sec:methods}.

We use discussions from Change My View (CMV) -- a unique forum (subreddit) on Reddit -- as an ideal test case due to its abundance with polarized discussions on multiple topics.  We carefully studied a few dozen discussions in developing the tagset and the annotation schema. Consequently, we have created a dataset of thousands of manually annotated utterances from hundreds of CMV discussions. A detailed description of the data and the annotation procedure is provided in Section \ref{sec:data}. The annotated dataset is shared with the community\footnote{Data, annotation guidelines and full annotation report can be found here <URL removed to maintain anonymity>.}.

{\bf Contribution} The contribution of this work is threefold: (i) We propose a novel framework (annotation scheme, tagset, task) for conversational discourse parsing, especially designed for non-convergent discussions, so common online, (ii) We share the first labeled conversational dataset annotated with the proposed scheme, and (iii) we demonstrate the learnability of the proposed tagset and provide an extensive analysis of the relevant features and the way different levels of discourse interact. 

\ignore{Second, as it is empirically evident that raising participants' awareness to the dynamics of the discussion in real time influences their discursive behavior during the discussion \cite{wise2011analyzing,yoon2011using}, we aim at producing appropriate intervention cues in real time. }

\ignore{The remainder of this paper is organized as follows: in Section \ref{sec:theory} we introduce the theoretical framework we build on and survey previous work.
The new discourse annotation scheme and the computational task are described in Section \ref{sec:annotation_scheme}. 
The data and the annotation protocol are described in Section \ref{sec:data}. We describe the experimental setting and discuss the results in Section \ref{sec:results}. }


\section{Related Work}
\label{sec:theory}
\subsection{Theoretical Framework}
\label{subsec:fruitful}
Oftentimes, a successful discussion is conceptualized in terms of convergence, an increment in the overlap of participants shared knowledge \cite{teasley2008cognitive}.  The convergence metaphor is viable for discussions that revolve around a problem or a task participants need to solve or achieve together, e.g. \cite{chiu2000group,diziol2010using}. It captures the gradual process through which a consensual answer is achieved \cite{lu2011collaborative}, a problem is resolved by a support agent \cite{bhuiyan2018don,oraby2017may} or collaborative learning in which students solve math problems \cite{schwarz2000two}.

Nonetheless, social media is replete with non-convergent discussions, for example discussions about morality, historical events, politically charged debates, and so forth.  Consensual answer is not expected in non-convergent discussions, yet these discussions may serve as a fruitful venue for the development of dialogic agency \cite{parker2006public}. Dialogic agency is related to the Bakhtinian concept of Internally Persuasive Discourse (IPD), a discursive regime, where participants examine and refine their vested truth -- ideas, consciousness, anthology -- in light of alternatives and critique provided by their peers, and even transcend beyond them \cite{bakhtin1981dialogic,bakhtin1986speech}. The quality of such non-convergent discussions, therefore, depends on participants’ \emph{responsiveness} -- their ability and will to consider their interlocutor’s point of view. Taking into account the interlocutor’s perspective does not necessarily imply agreement with her. Rather, it is about acknowledging the essence of the message conveyed in one’s response \cite{matusov2009bakhtin}.

Inspired by Bakhtin’s concept of dialogicity, we posit that a non-convergent discussion is productive as long as discussants exchange new ideas, while being responsive to alternatives and critiques offered by other interlocutors. The discourse annotation schema developed in Section \ref{sec:annotation_scheme} is designed to cover a range of discursive moves that reflect the speakers' responsiveness and the way it sets the trajectory of a discussion.   

\subsection{Conversational Discourse Analysis}
\label{subsec:DA}
Bakthin's dialogical stance has already been employed in the analysis of discourse in the field of collaborative learning, e.g., \cite{trausan2014polycafe,hennessy2016developing}. 
However, these applications were  not sensitive to non-productive interactions -- cases in which the discussion has deteriorated, which are common in online discourse. Recognizing  potential ``bad'' discursive moves is critical for improving the productivity of a discussion. Moreover, we conceptualize importance and productivity, in terms of \emph{potential} impact on the unfolding conversation. A comprehensive review of other annotation schemas discussions in educational settings can be found in \cite{noroozi2018promoting}. 
We are interested in a system (and a coding scheme) that captures the evolution of a discussion in an online manner -- allowing intervention when needed.

Zhang et al. \shortcite{zhang2018conversations} analyzed the role of politeness (or lack thereof) in
keeping conversations on track. Among other methods, they asked annotators to intuitively predict which conversation will get derailed, with emphasis on personal attacks and aggressiveness as causes. This work is the closest to ours, as it tackles early signs of deterioration in an online manner. However, Zhang et al. focus on politeness while we propose a comprehensive framework for discourse annotation and analysis. 
Another relevant work is that of \cite{cheng2017anyone}, who found that trolling behavior is not confined to a vocal and antisocial minority, but rather ordinary people can engage in such behavior as well. Such behavior is not merely due the individual’s mood, but also the surrounding context of the discussion.

Two important lines of work include general discourse parsing, e.g., \cite{core1997coding,jurafsky1997switchboard,stolcke2000dialogue,khanpour2016dialogue,ji2016latent} and argumentation mining, e.g., \cite{suthers2003representational,erkens2008automatic,rose2008analyzing,pinkwart2009evaluating,mclaren2010supporting,klebanov2016argumentation,bar2017stance}, among others. These works differ from ours in a number of fundamental ways. First, we are not interested in convergence or persuasion. Second, we are interested in capturing responsiveness in the context of IDP. That is, we are not interested in labeling the grammatical function of an utterance (such as 'a question'), nor merely the type of the argument, but rather to label how it relates to previous utterances (such as `a request for clarification of previous statement'). Our focus is on \emph{discursive moves} rather than on argument types\footnote{This is consistent with our interest in non-convergent discussions vs. dialogues striving for agreement or debates culminating in a `victory'.}.

\subsection{Non-conversational Discourse Analysis} 
\label{subsec:pdtb}
Discourse is also analyzed within a single textual unit\footnote{The terms `discourse analysis' and `discourse parsing' often used to refer to non-conversational units.}, modeling its coherence and topic drift  \cite{grimes1975thread,hobbs1985coherence,barzilay2004catching}. Recent work aims at modeling the relations between consecutive sentence pairs from the same document. The Penn Discourse Tree Bank (PDTB) 2.0 is composed of sentence pairs from five sections of the WSJ news stories, annotated with a set of discourse markers such as {\em but}, {\em and}, {\em as}, and {\em if} \cite{prasad2008penn}. State-of-the-art results are achieved using a bi-LSTM coupled with a BERT Transformer \cite{nie-etal-2019-dissent}.

Naturally, the two levels of discourse -- the in-utterance sentence-pair level and the conversational level, complement each other. \ignore{Moreover, the discourse markers within an utterance are the basic building blocks that serve the discursive strategy of a speaker in a conversation.} We therefore use the model proposed by Nie et al. \shortcite{nie-etal-2019-dissent} to enrich the representation of the conversational data by adding PDTB tags to each utterance.


\section{Discourse Annotation Scheme}
\label{sec:annotation_scheme}
We have developed a discourse annotation schema that captures the dynamics in a non-convergent discussion, with emphasis of speakers' responsiveness to previous utterances. This schema reflects the dynamics of an \emph{ongoing} discussion, lending itself to sequential annotation, in which an annotator labels each utterance as she reads it (and not in an ad-hoc manner).  

We define potentially\footnote{We refer to the \emph{potential} of the act since we predict its contribution in an online manner. At the time of the utterance the actual significance is not yet known.} productive discursive moves in light of Gricean principles \cite{grice1975logic}: (i) The interaction should revolve around the topic of discussion. Hence, discursive moves that shift the topic entirely are potentially unproductive, (ii) Utterances adding knowledge (or a different perspective) potentially contribute to the discussion and are likely to encourage the continuity of the conversation, whereas repetition may hasten the termination of the discussion or steer it into a loop, (iii) Mere addition of new knowledge, even if relevant and viable, is insufficient. Rather, the interlocutors are expected to be responsive to each other, taking into account the (essence of the) preceding utterances. Ignoring, distorting, practicing pettiness or referring only to insignificant elements of the previous utterance has the potential to harm the discussion, and (iv) Politeness and aggression impact the responsiveness of the conversants. Some of these maxims are echoed implicitly by \cite{barron2003smart,zhang2017characterizing,cheng2017anyone,zhang2018conversations}. 
\rev{These Grician principles serve in determining the level of responsiveness. We wish to stress that the term \emph{responsiveness} is used in the Bakhtinian sense of dialogic agency and IPD, as presented in Section \ref{subsec:fruitful}, and not as mere counts of utterance-response turn-taking.}

Guided by these principles, we thoroughly analyzed dozens of discussions from Reddit's CMV (see Section \ref{sec:data}), identifying significant discursive moves that reflect high and low levels of responsiveness. This qualitative analysis served as the basis for the new annotation scheme, introduced below (Section \ref{subsec:tagset}). The tagset was revised and refined through an initial annotation process, having a group of experienced meta-annotators\footnote{The meta-annotators have a published track record in qualitative discourse analysis. Details omitted in order to preserve anonymity.} discuss the annotation schema and the various discursive strategies observed.

\subsection{The Tagset}
\label{subsec:tagset}
The new tagset is presented in Table \ref{tab:tagset}, \rev{and  excerpts from two annotated conversations are presented in Table \ref{tab:example1} (examples for each of the tags are provided in Appendix A in the Supplementary Material).} The tags fall under four main categories,  corresponding to a range of (potentially) un/productive discursive moves. 

\begin{table*}[th!]
\centering\footnotesize
 \begin{tabular}{|m{0.7\linewidth}| c |} 
 \hline
  \rowcolor{Gray}  {\bf Description} & {\bf Tag}\\
 \hline\hline
 \rowcolor{Gray} 	1. Discursive moves that potentially promote the discussion \\
  \hline\hline
Moderating /regulating, e.g. ``let's go back to the discussion topic'' & \texttt{Moderation} \\ 
\hline
Request for clarification	 & \texttt{RequestClarification} \\ 
\hline
Attack on the validity of the argument (``can I see the evidence?'')	 & \texttt{AttackValidity} \\ 
\hline
Clarification of previous statement (utterance)	 & \texttt{Clarification} \\ 
\hline
Informative answer of a question asked (rather than clarifying )	 & \texttt{Answer} \\ 
\hline
A disagreement which is reasoned, a refutation. Can be accompanied by disagreement strategies & 	\texttt{CounterArgument} \\ 
\hline
Building/extending previous argument (w/o negation or disagreement). The speaker takes the idea of the previous speaker and extends it.	 & \texttt{Extension	} \\ 
\hline
A viable transformation of the discussion topic	 & \texttt{ViableTransformation} \\ 
\hline
Personal statement ``this happened to me'')	 & \texttt{Personal} \\ 
\hline \hline
\rowcolor{Gray}	2. Moves with low responsiveness \\
  \hline \hline
A severe case of low responsiveness, such as continuous squabbling and quarreling  & 	\texttt{BAD} \\ 
\hline
Repeating previous argument without any new detail or substantive variation	 & \texttt{Repetition} \\ 
\hline
Response to ancillary /distract the discussion	 & \texttt{NegTransformation} \\ 
\hline
Negation/disagreement without reasoning	 & \texttt{NoReasonDisagreement} \\ 
\hline
Convergence towards previous speaker	 & \texttt{Convergence	Agreement} \\ 
\hline
Announcement that the issue is not solvable yet there is legitimacy to the other’s voice	 & \texttt{AgreeToDisagree} \\ 
\hline \hline
\rowcolor{Gray}	3. Tone and style \\
  \hline \hline
\rowcolor{Gray}	3.1  Negative tone and style \\
  \hline\hline
Aggressive and Blatant (+ intensifying qualifiers) , e.g., ``this is stupid'' & 	\texttt{Aggressive} \\ 
\hline
Ridiculing the partner (or her argument)	& \texttt{Ridicule} \\ 
\hline
Addressing (complaining about) a negative approach e.g. ``you were rude to me'' & 	\texttt{Complaint} \\ 
\hline
Sarcasm/ cynicism /patronizing	 & \texttt{Sarcasm} \\ 
\hline\hline
\rowcolor{Gray}	3.2  Positive tone and style 	\\
  \hline\hline
Attempts to reduce tension: respectful, flattering , socially unifying tone	 & \texttt{Positive} \\ 
\hline
Weakening qualifiers e.g. ``I’m not an expert in this topic...''	 & \texttt{WQualifiers} \\ 
\hline\hline
\rowcolor{Gray}	4. Disagreement strategies (applied in addition to CounterArgument) \\
  \hline\hline
  \rowcolor{Gray}	4.1  Easing tension \\
  \hline\hline
Softening the blow; Saying something nice before or after conveying the disagreement	 & \texttt{Softening} \\ 
\hline
Partial agreement yet disagreement , e.g. ``I disagree only with one part of your text''	 & \texttt{AgreeBut} \\ 
\hline
Explicitly taking into account other participants’ voices	 & \texttt{DoubleVoicing} \\ 
\hline
Using an external source (a URL, academic reference) to support a claim	 & \texttt{Sources} \\ 
\hline\hline
\rowcolor{Gray}	4.2  Intensifying tension \\
  \hline\hline
Reframing or paraphrasing the previous comment in a way that changes the original meaning/sheds light on the limitations of the argument &  \texttt{RephraseAttack} \\ 
\hline
Critical question, phrasing the (counter) argument as a question	 & \texttt{CriticalQuestion} \\ 
\hline
Offering an alternative without directly refutation or disagreement & 	\texttt{Alternative} \\ 
\hline
Direct disagreement (“I disagree'', ``this is simply not true'')	 & \texttt{DirectNo} \\ 
\hline
Refutation focuses on the relevance of previous claim	 & \texttt{Irrelevance} \\ 
\hline
Breaking previous argument to pieces without real coherence & 	\texttt{Nitpicking} \\ 
\hline
\end{tabular}
 \caption{Tagset for non-convergent discourse parsing. The tags are divided to four discourse categories. \rev{Concrete examples for each of the tags are available in Appendix A in the supplementary Material}.}
  \label{tab:tagset}
\end{table*}

The first two categories capture \emph{responsiveness quality} -- knowledge exchange or the lack of it. For example, \texttt{CounterArgument}, \texttt{RequestClarification}, or \texttt{Extension} (of an argument made beforehand). 

The third category corresponds to \emph{style and tone}. We distinguish between positive (e.g., appreciation of the merit of one's response even if disagreeing with it, using humor, attempting to lessen the tension, predicting objections and tackling them), and negative attitudes (e.g. sarcasm, blaming the other interlocutor, expletive language). 

The forth category -- \emph{Disagreement Strategies} captures a finer grained counter argumentation.
\rev{The need for these nuanced subcategories is supported by a rich body of empirical literature, suggesting that different disagreement strategies intensify or soften aggression in argumentative discussions, e.g., \cite{felton2009deliberation,locher2010power,angouri2010you,lu2011collaborative,shum2013politeness,cheng2017anyone,cougnon2019mixed}.}
For example, consider the intuitive difference between the statement ``that's not true!'' (\texttt{DirectNo}) and a discursive move that `softens the blow' by preceding the counter argument with a compliment like ``that's a good point, but...'' (\texttt{Softening}) -- the former is potentially destructive, while the latter has the potential to ease tension. \rev{Other disagreement strategies, such as \texttt{Nitpicking} emerged as we qualitatively analyzed the initial data. Accordingly, the tags under category 4.1 reflect strategies that ease potential tension;  whereas those in category 4.2 reflect strategies that tend to intensify it.}

\subsection{Annotation Procedure}
\label{subsec:guidelines}
The unit of analysis is a full utterance. Annotators were asked to classify each utterance in terms of the central response made vis-a-vis the preceding utterances (and only the preceding utterances). Annotators were asked to code the essence of the utterance, considering its potential in shaping further utterances, rather than to dissect and annotate each and every part of the response. This gestalt viewpoint on responses is well established from a theoretical perspective \cite{gunawardena1997analysis}.

We allow multiple labels to be assigned to a single utterance. Tag collocation captures fine grained argumentation along with the interlocutor's tone and allows a responder to address a specific critique and extending a previous argument (see Table \ref{tab:example1} for examples).  

Although annotator agreement is not forced, annotators have a consolidation session after the annotation of each discussion. Consolidation was found to increase the quality of the annotation. See more on agreement in Section \ref{sec:data}.

\subsection{Examples: Labels and Discourse Trajectory}
\label{subsec:as_example}
We now provide a brief analysis of two excerpts from two longer discussions taken from Reddit's CMV subreddit \rev{(A detailed description of the CMV and the dataset is provided in Section \ref{sec:data})}. These examples demonstrate the effectiveness of the annotation schema in capturing (sequences) of un/productive discursive moves.

Table \ref{tab:example1} (top) presents a three-turn sequence extracted from a longer discussion about `racist words'. Speaker $A$'s discursive strategy (turn 7) is to ignore the sarcastic tone (\texttt{Sarcasm}) of speaker $B$ (turn 6), only providing the requested clarification (\texttt{Clarification}). Consequently, $B$ responds more positively, respectfully and on point by strategically adding positive tone (\texttt{Softening} and \texttt{WQualifiers}) to his main move -- \texttt{CounterArgument}. In essence, $A$'s choice to ignore $B$'s tone promotes the complex discursive move taken by $B$ to put the discussion back on track.

\begin{table*}[th!]
\centering\footnotesize
 \begin{tabular}{| c | c |>{\centering\arraybackslash}m{0.53\linewidth} |>{\centering\arraybackslash}m{3.7cm}|} 
 \hline
 \rowcolor{Gray}
  {\bf Turn} & {\bf Speaker} & {\bf Text} & {\bf Labels} \\ 
 \hline\hline
 6	& B & So why are we even discussing this right now if it's not your responsibility to change my mind. Shouldn't you be somewhere else not interfering? &	\texttt{RequestClarification}, \texttt{Sarcasm} \\ 
 \hline
7 &	A & I never said not interfering. I do not think it is worth my time trying to change racists. It is not worth my time to prevent them from speaking. It is worth my time to signal to others that the person ideas are racist. I am not going to change David Duke and I can't get him to shut up, by shutting his ideas I can help send the message that we do not tolerate his ideals and if you espouse them we will not tolerate you. & \texttt{Clarification} \\
\hline
8 &	B & There's a time and a place for outright blockading of the views of others if they can't find a healthy way to express them, but there are absolutely opportunities to help individuals progress towards changing their beliefs and behaviors. If there weren't then essentially it would be hopeless for gang members, sex offenders, or other criminals to ever change. There might be professionals out there who facilitate the change process best, but if it's impacting our society we absolutely have a responsibility to try and change those behaviors if confronted with them, in addition to preventing others from modeling them. &	\texttt{CounterArgument}, \texttt{Positive}, \texttt{Softening} \\
\hline\hline
\rowcolor{Gray}
& & & \\
\hline\hline
2 & A & You don't have to eat these shared group meals. If you voluntarily choose of your own free will to cook and eat a group lunch for a specific group of people then of course the meal should be something everyone in the group can eat. Otherwise you aren't cooking for a group, you're cooking for a portion of the group & \texttt{CounterArgument} \\ \hline
3 & B & ``something everyone in the group *can* eat'' Unless they're been bitten by a Lone Star Tick, there's nothing physically preventing them eating meat. & \texttt{NegTransformation}, \texttt{Ridicule} \\ \hline
4 & A &  Sure, if you want to be an abrasive aggressive asshole who doesn't exist within society but instead on the outskirts who hates people and is purposefully antagonist and rude, you could say that yes, it is physically possible for vegans to eat meat. But the rest of us here in society who aren't purposefully trying to be an asshole, understand that ``what everyone can eat'' isn't about literal physical ability. & \texttt{CounterArgument}, \texttt{Aggressive} \\ \hline
5 & B & `` `what everyone can eat' isn't about literal physical ability''. Then use your words: ``the meal should be something everyone in the group \textasciitilde can\textasciitilde *wants to* eat''. ``if you want to be an abrasive aggressive asshole'' You're the one flinging insults man. & \texttt{CounterArgument}, \texttt{Complaint} \\
\hline
 \end{tabular}
 \caption{Excerpts from two CMV discussions -- a productive one (top) and an unproductive discussion (bottom).}
  \label{tab:example1}
\end{table*}

Unfortunately, not all discussions recover. Iterating between \texttt{NegTransformation} ([petty] distraction), \texttt{Ridicule} and \texttt{Aggressive} signals that the conversation may not be going well (\texttt{BAD}). Table \ref{tab:example1} (bottom) presents an excerpt from a discussion about the ``appropriateness of bringing meat to a potluck with a vegetarian''. This sequence of discursive moves pushes the discussion from bad to worse, essentially deeming it unproductive -- for dozen or so discursive turns (not in the table) the dialogue is off topic, oscillating the semantic ``can'' vs. ``want'' argument, all embedded in insults.  \rev{This is the case in which a conversational discourse parser could be used to produce a gentle warning, alert a human moderator or prompt the participants with a suggestion for a different discursive strategy that will break the vicious cycle.} 

\rev{The two excerpts in Table \ref{tab:example1} provide an important observation -- the deterioration of a conversation is a process, reflected by a \emph{sequence} of moves.  Both cases started with a potentially ``bad''  move (\texttt{Sarcasm} and \texttt{Ridicule}), yet one recovered, whereas the other deteriorated. As the discussion on the top demonstrates, one bad move, one blatant, aggressive move, does not always bring about deterioration.}

\section{Data and Annotations}
\label{sec:data}
\paragraph{Reddit and r/ChangeMyView}  \ignore{Reddit is one of the most popular websites in the world, falling short behind Google and YouTube, with an average of over 300 million active users  per month\footnote{Reddit Now Has as Many Users as Twitter, and Far Higher Engagement Rates, \href{https://goo.gl/cLcMoY}{https://goo.gl/cLcMoY}, published 4/20/2018, retrieved 9/13/2020.}. Originally a link sharing platform, Reddit evolved into an active system of dedicated forums. }
Change My View, commonly known as CMV, is a unique forum in the Reddit platform. The premise of CMV is to promote a well reasoned discussion on contentious issues. Users initiate a discussion with a provocative statement, backing it by some initial argumentation, prompting the community to ``change the author's initial viewpoint''. Good argumentation is voted up and convincing arguments are awarded a `delta'. The community is strictly moderated thus trolling and ad-hominem attacks are deleted, ideally creating a sterile debate atmosphere. However, discussions often turn heated, presenting a heavy use of sarcasm and aggressive language in spite of the moderation.
CMV is, therefore, an ideal arena for stimulating open-ended and non-convergent discussions, rich enough in the number of users, variety topics discussed, argumentation styles and productivity of discussion.

\ignore{
\NegTransforamtion{ yes , here there are : example 1:  “Cool. Okay, so you're at least consistent. Lets keep pushing forward.”
example 2: I'm stubborn. It's often very difficult to tell when someone is arguing in bad faith, or just struggling to be charitable. I'm going to keep going as of this point. I think there's a chance we can get to some common ground.
example 3 : 
13. DawgDatsAGreatPost: >>> I did. As per Merriam-Webster: <quote> Gender: the state of being male or female.</quote> I only see two choices there. What? Merriam-Webster is "wrong" too?
14. EighthScofflaw: >>> Good, I'm glad you read the whole definition. <quote>(typically used with reference to social and cultural differences rather than biological ones)</quote> Good job.}
\NegTransforamtion {perhaps mention the work by zhang here ? zhang2017characterizing}
}

\paragraph{Discussions as trees and branches}
A discussion unfolds in a tree structure. Given a post, a user can reply directly to the post. Branching occurs if two or more users reply directly to the same post. Branching happens for reasons ranging from a naive chaining mistakes to conscious decisions to ``take the discussion to another direction''. Modeling the reasons for branching and their effect on the discussion is beyond the scope of this paper. For the sake of simplicity, in this paper we view a branch -- an ordered sequence of posts (root to leaf), as a discussion. Multiple branches of the same tree are considered separate discussions on the same topic, although some of them may share the branch prefix (all of them share the root -- the original prompt). Indeed, a manual examination of a sample of trees shows that it is reasonable to consider branches of the same tree as separate discussions if the shared prefix is not dominant (in length) compared to the mutual exclusive suffixes of the branches.

\paragraph{The Annotated Dataset} Sixteen thousands discussions that took place between April 2017--January 2018 were extracted from Reddit/r/CMV and converted to conversation trees. 1,946 branches from 100 trees were annotated by a group of trained annotators. Each discussion was annotated by two annotators. We consider annotations as valid only if both annotators agreed on a label. Agreement was reached in 95\% of the instances (lowest agreement for tag type was 91\%). 

The dataset contains a total of 10,559 posts out of which 9,620 were labeled with 17,964 tags (31 unique tags). \rev{Average number of users per tree, average length of a post and further details are provided in Table \ref{tab:treestats}. }

\begin{figure}[h]
\centering
\includegraphics[width=.9\linewidth]{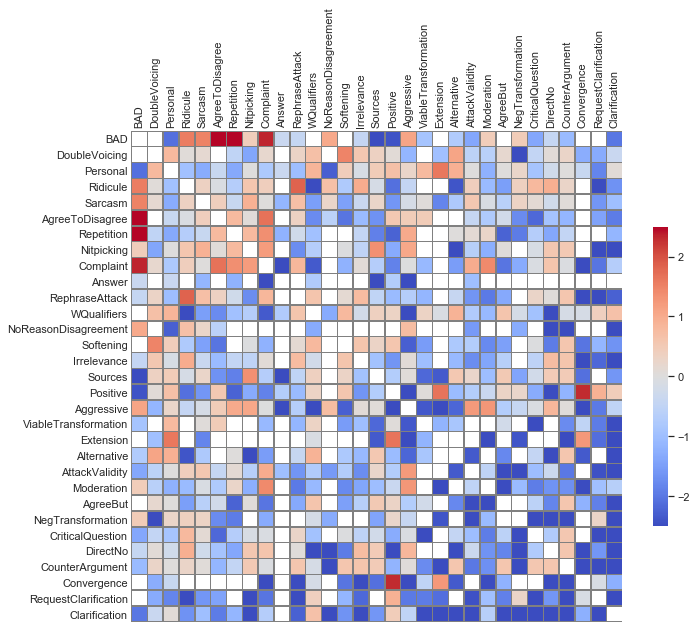}
\caption{\rev{Pointwise Mutual Information (PMI) for tag collocation.}}
\label{fig:pmi_tags}
\end{figure}

\begin{figure}[h]
\centering
\includegraphics[width=.9\linewidth]{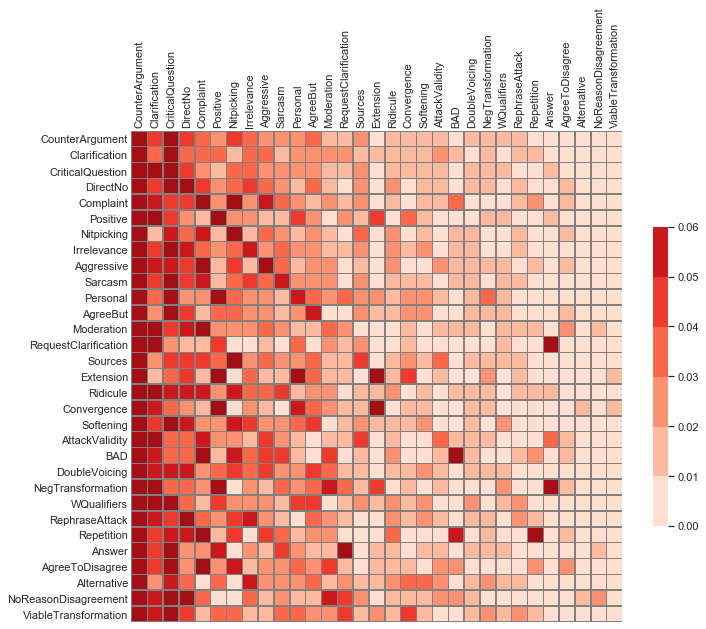}
\caption{\rev{Tag Transition probabilities.}}
\label{fig:transition_probs}
\end{figure}

\rev{Tag collocation measured through Point-wise Mutual Information (PMI) values is presented in Figure \ref{fig:pmi_tags}. \texttt{Ridicule} and \texttt{RephraseAttack} have high PMI, while \texttt{Ridicule} and \texttt{Clarification} have low PMI. Transition probabilities between tags in consecutive utterances are presented in Figure \ref{fig:transition_probs}. As expected, we observe that \texttt{RequestClarification} is relatively likely to be followed by \texttt{Clarification}, which in turn is likely to be followed by \texttt{CounterArgument} or \texttt{CriticalQuestion}.}

\rev{A careful examination of the collocations and the transition probabilities reveals intricate conversational dynamics, e.g., the surprising likelihood of \texttt{Extension} to follow \texttt{Convergence}, as if one discussant is not happy with the convergence of ideas and tries to keep the discussion going. The collocation and transition probabilities are leveraged as we model the discourse parsing task as a sequence model problem. The computational details are presented in the next section.}

\begin{table}[h!]
\centering
 \begin{tabular}{ r l r} 
 Variable & Value & STD. \\
 \hline
Num of trees    &	100  & -- \\
Total users	&   1610    & -- \\
Total branches  &	1946 &  --   \\   
Total nodes	&   10559    &  -- \\
Total labeled nodes	&   9620    &  --\\
Total labels for all nodes	&   17964    & --  \\
Total tokens & 1143777 & --\\
Avg. branches per tree  &	19.5  &  10.4 \\
Avg. nodes per tree &	105.6 &  60.8 \\
Avg. branch length  &	8.3 &  7.2  \\
Avg. tree depth    &	20.9  &  16.6 \\
Avg. users per branch   &	3.3  &  1.1 \\
Avg. users per tree &	22.8   & 12.4 \\
Avg. nodes per user &	6.4  &  11.9\\
Avg. tokens per post (node) & 108.3 & 139.9\\
\hline
 \end{tabular}
 \caption{\rev{Dataset statistics.}}
 \label{tab:treestats}
\end{table}

\section{Computational Approach}
\label{sec:methods}

\subsection{Task Definition}
\label{subsec:task}
\paragraph{Formal Definitions} The discourse parsing task can be broadly defined in the following manner: find the sequence $ L$ that maximizes $P(L|T)$ where $L = l_{1},...,l_{n}$ denotes a sequence of labels, $l_i \in \mathcal{L}$, a predefined tagset. The sequence $L$ corresponds to a sequence of texts/utterances $T = t_{1},...,t_{n}$. 

We note that $L$ actually encode a sequences of \emph{sets} of labels ($l_i = \{l\}_1^k \subset \mathcal{L}$) as each utterance can be assigned more than one label (e.g. $<$\emph{CounterArgument, Positive, Softening}$>$ -- its discursive move, tone, disagreement strategy, see Turn 8 (line 3) in Table \ref{tab:example1}, top). 

\paragraph{Naive Classification} The most straight forward approach is to apply supervised learning, using each textual unit (utterance) as the basic unit for label prediction. Utterances like ``My view is that this is all nonsense'', ``but that's just my two cents''  or ``I disagree'', intuitively suggest that discourse markers are part of the text and could be learned efficiently. In other words, we want to maximize $P(L|T)$ under the naive assumption that the order of the texts in $T$ is insignificant: $P(L|T) = \Pi_{t_{i=1...|T|} \in T} P(l_{i}|t_{i})$. A similar approach is considered by \cite{hirschberg1993empirical}.

\paragraph{Sequence Labeling:} Dialogical acts tend to correspond, at least in a cooperative conversation (and under Gricean framework described in Section \ref{sec:annotation_scheme}). Intuitively, a question is likely to draw an answer and an aggressive comment is likely to draw an aggressive response.
This assumed conversational structure, \rev{supported by the collocation and transition probabilities presented in Figure \ref{fig:pmi_tags} and \ref{fig:transition_probs}}, suggests looking at the annotation task as a structured prediction problem, in which labels are not assigned based on single textual units but depend on other labels in the sequence. The general formulation of the annotation-optimization task is to maximize $L$ in $P(L|T) = \Pi_{t_{i=1...|T|} \in T} P(l_{i}|t_{i}, L^{-i})$, where $L^{-i}$ denotes the labels predicted in the sequence except the $i$-th label. 

\ignore{This can be done by addressing the annotation task using methods such as Hidden Markov Models \cite{stolcke2000dialogue,collins2002discriminative}, Conditional Random Fields  \cite{lafferty2001conditional,bakir2007predicting,smith2011synthesis} or  with deep neural architectures \cite{bengio2015scheduled,zhang_neural_2015,joty2018modeling}. With the exception of Stolke et al.\shortcite{stolcke2000dialogue} these works present the different frameworks for structured prediction, either dealing with the general algorithmic framework or lower level linguistic tasks like POS tagging and grammatical parsing. The seminal work by Stolke et al. proposes using n-gram discourse models trained with an HMM. }

\ignore{\paragraph{Relaxing Conditions} In this paper we aim at demonstrating the potential of the newly introduced annotation scheme and the learnability of the tagset. As this is still a work in progress, we do not yet have enough annotated data thus the use of proper sequence prediction algorithms (e.g. CRF, LSTM) is beyond the scope of this paper. Instead, we use standard classifiers and introduce sequence and collocation properties as features in an ad-hoc manner, as explained below.  \oren{I think we should remove this paragraph. it draws attention. }}

\subsection{Feature Types and Representation}
\label{subsec:features}

In order to assign labels to texts we stack a number of binary classifiers - training a model for each label. Each classifier learns a model for a single label. 

The input text is represented by a combination of features. Feature types are presented below and results are reported in Section \ref{sec:results}. 

\paragraph{Bag of Words (BOW)} The most basic way to represent a piece of text is by the words it contains. We considered different dimensions for the BOW vectors: 300, 500, 1000 \& 1500. We experimented with both a TF-IDF weighing \cite{salton1988term} and standard binary vectors of the most frequent words.

\paragraph{Embeddings: GloVe, Doc2vec} Two types of  embeddings were considered: a 300 dimension GloVe word representations \cite{pennington2014glove}, averaged for all the words in a post, and a another vector created by doc2vec \cite{le2014distributed}, instead of averaging.

\paragraph{LIWC Categories} The Linguistic Inquiry and Word Counts (LIWC) dictionary is used to assign words to cognitive and emotion categories \cite{pennebaker2001linguistic}. We expect open-ended argumentative discourse to be emotionally charged. Utterances are represented by a vector of LIWC categories -- each entry reflects the weight of the corresponding LIWC category in that text. 

\paragraph{PDTB discourse relations} While this work is focused on conversational structure, the discursive structure within a single document or utterance is also of interest (see Section \ref{subsec:pdtb}). We hypothesize that the inner structure complements the conversational structure, therefore can serve as a powerful signal in modeling the latter. We use a Bi-LSTM fine-tuned over BERT as described in \cite{nie-etal-2019-dissent} to label the inner structure (semantic relation between each pair of consecutive sentences) of each post with the PDTB2.0 tags \cite{prasad2008penn}. We then represent each post by two vectors: one of the counts of the PDTB tags found in the post, and the other of the counts of bigrams of PDTB tags. These vectors are used as an additional layer of features representing a post. 

\paragraph{Conversational Sequence} The main objective in the defined discourse parsing task is to model dialogicity. A text $t_i$ is a direct response to a previous comment $t_{i-1}$ and may also be effected (argumentatively, stylistically, emotionally) by previous comments $t_{1-2}, t_{i-3} ...$ made by one or more interlocutors. We therefore expect the text of previous comments to carry some relevant signal. We experiment with history length $\in \{0,1,2,3\}$. 

\paragraph{Label Sequence} In essence, discourse parsing is a sequence prediction task as the labels of the current text are expected to be related to previous labels. Developing a sequence prediction of multi-class-multi tag is beyond the scope of this paper. However, in order to demonstrate the learnability of the proposed tagset we take the gold standard labels of previous texts as given. Labels of a previous text $t'$ are represented as a binary vector in the dimension of the tagset. We experiment with using the labels of the 1,2 \& 3 previous posts, as well as no previous labels at all. 

\paragraph{Label Collocation} The proposed annotation scheme allows a text to be labeled with more than a single label (e.g. labels for function, tone and argumentation strategy, see Section \ref{subsec:tagset}). Label collocation is, therefore, similar to the sequential nature of the task -- a specific label may be more likely to appear with other labels. We consider the use of the collocated labels as yet another feature type. 

\paragraph{Combining Feature Types} Each feature type is a vector. Combining feature types is done by concatenating the respective vectors. For example, when using features of previous posts (either BOW, GloVe, LIWC or labels) the vectors of the texts are simply appended. That is, if $\overrightarrow{v_{i}}$ and $\overrightarrow{v_{i-1}}$ are the GloVe vectors representing texts $t_{i}$ and $t_{i-1}$ respectively, and we only use the text of the previous post $t_{i-1}$ along with the text of the current post $t_{i}$ for predicting the label/s of $t_i$, the feature vector representation of $t_i$ is $\overrightarrow{v_i.v_{i-1}}$.

\subsection{Classification Models}
\label{subsec:classifiers}
\rev{We experimented with the following classifiers: Logistic Regression, Naive Bayes, Decision Trees, SVM, Feed Forward Neural Network, and a BERT Transformer\footnote{We used the sklearn package \cite{scikit-learn} for the LR, SVM, NB and DT algorithms and Roberta \cite{liu2019roberta} for the BERT transformer.}}.

\rev{The input for all classifiers, except the BERT Transformer, was a feature vector composed of a concatenation of some or all the feature categories described above.
The BERT transformer, on the other hand, takes a textual sequence as an input, thus the the feature vectors we used for other classifiers could not be used. The transformer was fed by the text of the current post. Since the input sequence of the BERT Transformer is limited to 512 tokens we opted for not using the text of preceding posts. Similarly, the text of the current post was truncated if it exceeded the allowed length. However, only $\sim$2.4\% of the posts exceeded that length.
We did manipulate the BERT input in order to introduce label sequence and collocation. This was done by appending the text of each post to ``dummy'' sentences listing the collocated and preceding tags in a simplified grammar\footnote{Input structure example: `previous tags: tag $t_1$ tag $t_2$ tag $t_3$, current post: <text of the current post>'. White spaces were added to tags composed of multiple words in order to have match the tag to the BERT dictionary (e.g., \texttt{CounterArgument} $\rightarrow$ 'counter argument').}.}

We note that a bi-directional neural architecture could not be used due to the ``online'' definition of the task -- annotating the discussion as it unfolds, rather than annotating a full discussion.


\begin{table}[h!]
\centering\footnotesize
 \begin{tabular}{ c  c  c  c  c } 
 \hline
 \rowcolor{Gray}
 {\bf Tag} & {\bf Prec} & {\bf Rec} & {\bf F1-score} & {\bf Prior}  \\ 
 \hline\hline
 \rowcolor{Gray}
  1. Promoting discussion \\
 \hline\hline
CounterArgument & 0.927 & 0.951 & 0.939 & 0.585 \\ 
Clarification & 0.872 & 0.768 & 0.817 & 0.100 \\ 
RequestClarification & 0.831 & 0.653 & 0.731 & 0.035 \\ 
Extension & 0.517 & 0.585 & 0.549 & 0.021 \\ 
Answer & 0.857 & 0.375 & 0.522 & 0.013 \\ 
AttackValidity & 0.500 & 0.520 & 0.510 & 0.026 \\ 
Moderation & 0.739 & 0.293 & 0.420 & 0.033 \\ 
Personal & 0.304 & 0.570 & 0.396 & 0.043 \\ 
ViableTransformation & 0.089 & 0.733 & 0.158 & 0.009	\\ \hline\hline
macro avg. & 0.626	& 0.606 &	0.560 &	0.096 \\
w.avg & \textbf{0.846} & \textbf{0.842} & \textbf{0.833} & \textbf{0.414}\\
\hline\hline
 \rowcolor{Gray}
  2. Low responsiveness \\ \hline\hline
Convergence & 0.632 & 0.511 & 0.565 & 0.026 \\ 
NegTransformation & 0.722 & 0.283 & 0.406 & 0.022 \\ 
NoReasonDisagreement & 0.455 & 0.357 & 0.400 & 0.010 \\ 
AgreeToDisagree & 0.131 & 0.423 & 0.200 & 0.013 \\ 
Repetition & 0.313 & 0.109 & 0.161 & 0.015 \\ 
BAD & 0.066 & 0.419 & 0.114 & 0.017 \\ \hline\hline
macro avg. & 0.386 & 0.350 & 0.308 & 0.017 \\ 
w. avg. & \bf0.429 & \bf0.362 & \bf0.335 & \bf0.019\\
\hline\hline
 \rowcolor{Gray}
  3. Tone \& style \\ \hline\hline
Complaint & 0.300 & 0.400 & 0.343 & 0.059 \\ 
Positive & 0.325 & 0.348 & 0.336 & 0.053 \\ 
Aggressive & 0.133 & 0.234 & 0.170 & 0.047 \\ 
Sarcasm & 0.123 & 0.243 & 0.164 & 0.044 \\ 
WQualifiers & 0.071 & 0.345 & 0.118 & 0.022 \\ 
Ridicule & 0.061 & 0.580 & 0.110 & 0.027 \\ \hline\hline
macro avg & 0.169 & 0.358 & 0.207 & 0.042 \\ 
w. avg & \bf0.198 & \bf0.345 & \bf0.233 & \bf0.046\\
\hline\hline
 \rowcolor{Gray}
  4. Disagreement strategies \\ \hline\hline
  \rowcolor{Gray} 
  4.1 Easing tension \\ \hline\hline
Sources & 0.884 & 0.884 & 0.884 & 0.042 \\ 
Softening & 0.330 & 0.446 & 0.379 & 0.053 \\ 
DoubleVoicing & 0.106 & 0.565 & 0.179 & 0.027 \\ 
AgreeBut & 0.056 & 0.867 & 0.106 & 0.024 \\ \hline\hline
macro avg & 0.344 & 0.690 & 0.387 & 0.036 \\ 
w. avg & \bf0.403 & \bf0.663 & \bf0.442 & \bf0.040\\
\hline\hline
  \rowcolor{Gray} 
  4.2 Intensifying tension \\ \hline\hline
  Nitpicking & 0.730 & 0.860 & 0.790 & 0.056 \\ 
CriticalQuestion & 0.786 & 0.668 & 0.722 & 0.118 \\ 
DirectNo & 0.260 & 0.258 & 0.259 & 0.073 \\ 
Irrelevance & 0.116 & 0.333 & 0.172 & 0.054 \\ 
Alternative & 0.082 & 0.366 & 0.133 & 0.017 \\ 
RephraseAttack & 0.042 & 0.514 & 0.077 & 0.020 \\ \hline\hline
macro avg & 0.336 & 0.500 & 0.359 & 0.056 \\ 
w. avg & \bf0.476 & \bf0.533 & \bf0.477 & \bf0.077 \\ \hline\hline
  \rowcolor{Gray}
  \bf{Total} \\
macro avg. & 0.399 & 0.499 & 0.382 & 0.055 \\ 
w.avg.	& {\bf	0.614}	&	{\bf	0.663}	&	{\bf	0.610} & {\bf0.237} \\ \hline
\end{tabular}
\caption{Best results for each tag grouped by categories (each group ordered by F1-score), allowing a different model for each tag. Macro average given for reference.}
\label{tab:detailedResults}
\end{table}

\begin{table}[h!]
\centering\footnotesize
 \begin{tabular}{p{0.015\textwidth}|p{0.01\textwidth}|p{0.01\textwidth}|p{0.01\textwidth}|p{0.025\textwidth}|p{0.025\textwidth}|c|c|c|c}
       \hline
 \rowcolor{Gray}
 {\tiny $T^C$} & {\tiny $T_2$} & {\tiny $L_1$} & {\tiny $B_1$} & {\tiny $PD^B_1$} & { \tiny $PD^U_2$} & {\bf w.P} & {\bf w.R} & {\bf w.F1} & {\bf m.F1} \\ 
 \hline\hline
     & & & & \cmark & \cmark & 0.281 &	0.233 &	0.253  & 0.061 \\ 
     & & \cmark & & & & 0.278    &	0.538   &	0.314 & 0.119   \\ 
     & & & \cmark & & & 0.387	&	0.356	&	0.367 & 0.181	\\ 
     & & \cmark & \cmark & & & 0.397	&	0.360	&	0.374 & 0.183	\\ 
     & & \cmark & \cmark & \cmark & &	0.388	&	0.353	&	0.366 & 0.181	\\ 
     & & \cmark & \cmark & \cmark & \cmark &  0.394 &  0.366 & 0.376 & 0.185 \\
      & \cmark & &   & &   & 0.311    &	0.260   &	0.279  & 0.083 \\
     & \cmark & \cmark &   & &   &  0.306 &	0.495   &	0.348 & 0.146 \\
     & \cmark & \cmark &   & \cmark & & 0.324    &	0.443   &	0.354  & 0.157\\
     & \cmark &  \cmark &   &  \cmark & \cmark  & 0.330   &	0.412   &	0.354  & 0.153 \\
     & \cmark &  \cmark & \cmark &  &  & 0.404 &	0.381   &	0.389  & 0.190 \\
     & \cmark &  \cmark & \cmark & & \cmark &  0.414    &	0.390   &	0.398  & 0.200 \\
     & \cmark &  \cmark & \cmark & \cmark & &   0.405    &	0.375   &	0.387 & 0.186 \\
     & \cmark &  \cmark & \cmark & \cmark & \cmark & 0.416   &	0.390   &	0.398 & 0.195 \\
     \cmark & \cmark & & & & &  0.474 &  0.438 & 0.452  & 0.235\\
     \cmark & \cmark & \cmark & & & & 0.439 &  {\bf 0.568} & 0.474  & 0.265\\ 
     \cmark & \cmark & \cmark & & & \cmark & 0.454	&	0.536	&	0.476	& 0.273\\ 
     \cmark & \cmark & \cmark & & \cmark & \cmark & 0.473	&	0.529	&	0.488 & 0.275	\\
     \cmark & \cmark & \cmark & \cmark & & & 	0.549	&	0.500	&	0.517 & 0.292 \\
     \cmark & \cmark & \cmark & \cmark & & \cmark & 	0.553	&	0.501	&	0.519	& 0.287\\
     \cmark & \cmark & \cmark & \cmark & \cmark & \cmark &   0.551 &  0.499 & 0.518  & 0.283\\
     \cmark & \cmark & \cmark & \cmark & \cmark & &    {\bf 0.555} &  0.509 & {\bf 0.526} & {\bf 0.297} \\
     \hline
 \end{tabular}
  \caption{Results of ablation test for average (weighted) over all labels. Macro F1 given for reference. Notation: $B$: BOW; $L$: LIWC; $PD$: PDTB labels; $PD^B$: bigrams of PDTB labels; Subscript index denotes the context length, {\em including} current post. $T^C$: tag collocation of current post. Note: $T_j$: context length, current post {\em not} included (note the exception).}
 \label{tab:ablation2}
\end{table}

\section{Results and Analysis}
\label{sec:results}
\rev{A set of 15 trees was set aside for testing. This test set is composed of 321 discussion branches with a total of 1835 nodes annotated with 3412 labels. 
Training was done in a 5-fold cross validation manner on the other 85 trees. This procedure was repeated in multiple settings differing in the classification model and combinations of features types used. For the sake of brevity we report the best result for each tag and category (Table \ref{tab:detailedResults}) and ablation results for the single model that performed best averaged over all tags (Table \ref{tab:ablation2}). All results are reported on the 15 trees in the held-out set.}

\paragraph{Model per tag:} \rev{We achieved a weighted F-score for all 31 tags is 0.61. The F-score for each individual tags varies from 0.939 (\texttt{CounterArgument}, prior 0.585) to 0.077 (\texttt{RephraseAttack}, prior 0.02). The average F-score for the first category (Promoting Discussion) is 0.833, significantly higher than the other categories. While \texttt{CounterArgument} has the highest prior (0.585), the high F-score for the first category cannot be attributed to higher priors as tags with low priors ranging from 0.1 to 0.013 achieved impressive F-scores ranging from 0.817 to 0.522 (e.g., \texttt{Clarification}, \texttt{Extension}, and \texttt{Answer}). A significant improvement over the prior was achieved for all the tags.}

\rev{Analysing the performance of the different classifiers we find that the BERT classifier, adopted for label sequence and collocation, achieved nine of the top results in Table \ref{tab:detailedResults}, including the impressive F-scores for \texttt{CounterArgument}, \texttt{Sources},\texttt{Clarification},  \texttt{Nitpicking}, and \texttt{RequestClarification}, most of which fall under Category 1 (Promoting Discussion). Surprisingly though, the BERT classifier performed poorly on all other tags. Further analysis of the variance of the BERT performance is beyond the scope of this paper and will be addressed in future work.}
\rev{Top results for the other tags were obtained using a different classifier and feature set for each tag. Each of the feature types listed in Section \ref{subsec:features} proved useful in achieving top results for at least one of the tags, with BOW, LIWC, label sequence and collocation being the most stable features across labels and classifiers.}

\paragraph{Single model for all tags:} \rev{Allowing only a single classification model for all tags, we achieve a weighted F-score of 0.526, using a  feed-forward network with three hidden layers, using a tanh activation and L2 regularization with min-max scaling. Nevertheless, while using a single model results in a significant drop of the weighted F-score, it is still impressive, given the number of classes (31) and the complexity of the task. Using a single model allows us to explore the contribution of the different feature types to the parsing task. Some ablation results are reported in Table \ref{tab:ablation2}. For example, using only the text (BOW) of the current post ($B_1$) achieves an F-score of 0.367, while using both the text and the LIWC features of the current post ($B_1$ and $L_1$) yields a small improvement to 0.374. 
Top results for a single classifier were achieved by using a feature vector combining the BOW ($B_1$), LIWC ($L_1$), PDTB discourse bi-grams ($PD^{B}_1$), collocation of tags in the current post ($T^C$), and the tag sequence of the two preceding posts ($T_2$).}

\paragraph{Noisy sequences and collocations:} \rev{In order to simulate a sub-optimal prediction of the preceding sequence we performed a set of experiments in which we introduce noise to the input by masking or randomly replacing 10\%, 20\% or 50\% of the collocated and preceding tags, or by adding tags at random (by priors). An F-score of 0.463 was achieved upon a random substitution of 50\% of the tags -- the largest drop observed. Masking only 10\% of the tags achieved an F-score of 0.523, and randomly replacing 10\% of the tags resulted in an F-score of 0.518, comparing to the 0.526 achieved without distorting the input.}

The results and analysis presented here demonstrate the learnability of the complex and nuanced schema we proposed. Moreover, it supports the theoretical framework of Dialogic Agency, IPD, and Responsiveness, presented in Section \ref{sec:theory}, thus promoting further research of conversational dynamics.

\section{Conclusion}
\label{sec:closure}
Many online discussions are non-convergent. We have argued that the productivity of a discussion should not be measured through convergence and that a new discourse annotation scheme is required in order to capture the productivity of online discussions. Inspired by theory of dialogism, we have proposed a new discourse annotation schema that can be used for early detection of the trajectories a discussion takes, and ultimately improve discussion quality. We have explored a number of classification algorithms and an array of feature types, demonstrating the learnability of the proposed schema. Making the annotated data public will allow the research community to further explore conversational dynamics -- a much needed undertaking in this era of highly polarized contentious discourse. 

\bibliography{CDP_ICWSM_21}

\begin{thebibliography}{}

\bibitem[\protect\citeauthoryear{Angouri and Tseliga}{2010}]{angouri2010you}
Angouri, J., and Tseliga, T.
\newblock 2010.
\newblock “you have no idea what you are talking about!” from
  e-disagreement to e-impoliteness in two online fora.
\newblock {\em Journal of Politeness Research} 6(1):57--82.

\bibitem[\protect\citeauthoryear{Bakhtin, Holquist, and
  Emerson}{1986}]{bakhtin1986speech}
Bakhtin, M.~M.; Holquist, M.; and Emerson, C.
\newblock 1986.
\newblock Speech genres and other late essays.

\bibitem[\protect\citeauthoryear{Bakhtin}{1981}]{bakhtin1981dialogic}
Bakhtin, M.~M.
\newblock 1981.
\newblock The dialogic imagination: Four essays by mm bakhtin (m. holquist,
  ed.; c. emerson \& m. holquist, trans.).

\bibitem[\protect\citeauthoryear{Bar-Haim \bgroup et al\mbox.\egroup
  }{2017}]{bar2017stance}
Bar-Haim, R.; Bhattacharya, I.; Dinuzzo, F.; Saha, A.; and Slonim, N.
\newblock 2017.
\newblock Stance classification of context-dependent claims.
\newblock In {\em Proceedings of the 15th Conference of the European Chapter of
  the Association for Computational Linguistics: Volume 1, Long Papers},
  251--261.

\bibitem[\protect\citeauthoryear{Barron}{2003}]{barron2003smart}
Barron, B.
\newblock 2003.
\newblock When smart groups fail.
\newblock {\em The journal of the learning sciences} 12(3):307--359.

\bibitem[\protect\citeauthoryear{Barzilay and Lee}{2004}]{barzilay2004catching}
Barzilay, R., and Lee, L.
\newblock 2004.
\newblock Catching the drift: Probabilistic content models, with applications
  to generation and summarization.
\newblock In {\em Proceedings of the Human Language Technology Conference of
  the North American Chapter of the Association for Computational Linguistics:
  HLT-NAACL 2004}.

\bibitem[\protect\citeauthoryear{Bhuiyan \bgroup et al\mbox.\egroup
  }{2018}]{bhuiyan2018don}
Bhuiyan, M.; Misra, A.; Tripathy, S.; Mahmud, J.; and Akkiraju, R.
\newblock 2018.
\newblock Don't get lost in negation: An effective negation handled dialogue
  acts prediction algorithm for twitter customer service conversations.
\newblock {\em arXiv preprint arXiv:1807.06107}.

\bibitem[\protect\citeauthoryear{Chen \bgroup et al\mbox.\egroup
  }{2018}]{chen2018fostering}
Chen, B.; Chang, Y.-H.; Ouyang, F.; and Zhou, W.
\newblock 2018.
\newblock Fostering student engagement in online discussion through social
  learning analytics.
\newblock {\em The Internet and Higher Education} 37:21--30.

\bibitem[\protect\citeauthoryear{Cheng \bgroup et al\mbox.\egroup
  }{2017}]{cheng2017anyone}
Cheng, J.; Bernstein, M.; Danescu-Niculescu-Mizil, C.; and Leskovec, J.
\newblock 2017.
\newblock Anyone can become a troll: Causes of trolling behavior in online
  discussions.
\newblock In {\em Proceedings of the 2017 ACM conference on computer supported
  cooperative work and social computing},  1217--1230.

\bibitem[\protect\citeauthoryear{Chiu}{2000}]{chiu2000group}
Chiu, M.~M.
\newblock 2000.
\newblock Group problem-solving processes: Social interactions andindividual
  actions.
\newblock {\em Journal for the theory of social behaviour} 30(1):26--49.

\bibitem[\protect\citeauthoryear{Core and Allen}{1997}]{core1997coding}
Core, M.~G., and Allen, J.
\newblock 1997.
\newblock Coding dialogs with the damsl annotation scheme.
\newblock In {\em AAAI fall symposium on communicative action in humans and
  machines}, volume~56.
\newblock Boston, MA.

\bibitem[\protect\citeauthoryear{Cougnon, Coppin, and
  Figueroa}{2019}]{cougnon2019mixed}
Cougnon, L.-A.; Coppin, J.; and Figueroa, V.~G.
\newblock 2019.
\newblock A mixed quantitative-qualitative approach to disagreement in online
  news comments on social networking sites.
\newblock {\em Social Media Corpora for the Humanities (CMC-Corpora2019)} ~32.

\bibitem[\protect\citeauthoryear{Diziol \bgroup et al\mbox.\egroup
  }{2010}]{diziol2010using}
Diziol, D.; Walker, E.; Rummel, N.; and Koedinger, K.~R.
\newblock 2010.
\newblock Using intelligent tutor technology to implement adaptive support for
  student collaboration.
\newblock {\em Educational Psychology Review} 22(1):89--102.

\bibitem[\protect\citeauthoryear{Erkens and
  Janssen}{2008}]{erkens2008automatic}
Erkens, G., and Janssen, J.
\newblock 2008.
\newblock Automatic coding of dialogue acts in collaboration protocols.
\newblock {\em International journal of computer-supported collaborative
  learning} 3(4):447--470.

\bibitem[\protect\citeauthoryear{Felton, Garcia-Mila, and
  Gilabert}{2009}]{felton2009deliberation}
Felton, M.; Garcia-Mila, M.; and Gilabert, S.
\newblock 2009.
\newblock Deliberation versus dispute: The impact of argumentative discourse
  goals on learning and reasoning in the science classroom.
\newblock {\em Informal Logic} 29(4):417--446.

\bibitem[\protect\citeauthoryear{Grice}{1975}]{grice1975logic}
Grice, H.~P.
\newblock 1975.
\newblock Logic and conversation.
\newblock {\em 1975}  41--58.

\bibitem[\protect\citeauthoryear{Grimes}{1975}]{grimes1975thread}
Grimes, J.~E.
\newblock 1975.
\newblock {\em The thread of discourse}, volume 207.
\newblock Walter de Gruyter.

\bibitem[\protect\citeauthoryear{Gunawardena, Lowe, and
  Anderson}{1997}]{gunawardena1997analysis}
Gunawardena, C.~N.; Lowe, C.~A.; and Anderson, T.
\newblock 1997.
\newblock Analysis of a global online debate and the development of an
  interaction analysis model for examining social construction of knowledge in
  computer conferencing.
\newblock {\em Journal of educational computing research} 17(4):397--431.

\bibitem[\protect\citeauthoryear{Hennessy \bgroup et al\mbox.\egroup
  }{2016}]{hennessy2016developing}
Hennessy, S.; Rojas-Drummond, S.; Higham, R.; M{\'a}rquez, A.~M.; Maine, F.;
  R{\'\i}os, R.~M.; Garc{\'\i}a-Carri{\'o}n, R.; Torreblanca, O.; and Barrera,
  M.~J.
\newblock 2016.
\newblock Developing a coding scheme for analysing classroom dialogue across
  educational contexts.
\newblock {\em Learning, Culture and Social Interaction} 9:16--44.

\bibitem[\protect\citeauthoryear{Hirschberg and
  Litman}{1993}]{hirschberg1993empirical}
Hirschberg, J., and Litman, D.
\newblock 1993.
\newblock Empirical studies on the disambiguation of cue phrases.
\newblock {\em Computational linguistics} 19(3):501--530.

\bibitem[\protect\citeauthoryear{Hobbs}{1985}]{hobbs1985coherence}
Hobbs, J.~R.
\newblock 1985.
\newblock On the coherence and structure of discourse.

\bibitem[\protect\citeauthoryear{Ji, Haffari, and
  Eisenstein}{2016}]{ji2016latent}
Ji, Y.; Haffari, G.; and Eisenstein, J.
\newblock 2016.
\newblock A latent variable recurrent neural network for discourse relation
  language models.
\newblock In {\em Proceedings of NAACL-HLT},  332--342.

\bibitem[\protect\citeauthoryear{Jurafsky, Shriberg, and
  Biasca}{1997}]{jurafsky1997switchboard}
Jurafsky, D.; Shriberg, E.; and Biasca, D.
\newblock 1997.
\newblock Switchboard-damsl labeling project coder’s manual.
\newblock {\em Technick{\'a} Zpr{\'a}va}  97--02.

\bibitem[\protect\citeauthoryear{Khanpour, Guntakandla, and
  Nielsen}{2016}]{khanpour2016dialogue}
Khanpour, H.; Guntakandla, N.; and Nielsen, R.
\newblock 2016.
\newblock Dialogue act classification in domain-independent conversations using
  a deep recurrent neural network.
\newblock In {\em Proceedings of COLING 2016, the 26th International Conference
  on Computational Linguistics: Technical Papers},  2012--2021.

\bibitem[\protect\citeauthoryear{Klebanov \bgroup et al\mbox.\egroup
  }{2016}]{klebanov2016argumentation}
Klebanov, B.~B.; Stab, C.; Burstein, J.; Song, Y.; Gyawali, B.; and Gurevych,
  I.
\newblock 2016.
\newblock Argumentation: Content, structure, and relationship with essay
  quality.
\newblock In {\em Proceedings of the Third Workshop on Argument Mining
  (ArgMining2016)},  70--75.

\bibitem[\protect\citeauthoryear{Le and Mikolov}{2014}]{le2014distributed}
Le, Q., and Mikolov, T.
\newblock 2014.
\newblock Distributed representations of sentences and documents.
\newblock In {\em International conference on machine learning},  1188--1196.

\bibitem[\protect\citeauthoryear{Liu \bgroup et al\mbox.\egroup
  }{2019}]{liu2019roberta}
Liu, Y.; Ott, M.; Goyal, N.; Du, J.; Joshi, M.; Chen, D.; Levy, O.; Lewis, M.;
  Zettlemoyer, L.; and Stoyanov, V.
\newblock 2019.
\newblock Roberta: A robustly optimized bert pretraining approach.
\newblock {\em arXiv preprint arXiv:1907.11692}.

\bibitem[\protect\citeauthoryear{Locher}{2010}]{locher2010power}
Locher, M.~A.
\newblock 2010.
\newblock {\em Power and politeness in action: Disagreements in oral
  communication}, volume~12.
\newblock Walter de Gruyter.

\bibitem[\protect\citeauthoryear{Lu, Chiu, and Law}{2011}]{lu2011collaborative}
Lu, J.; Chiu, M.~M.; and Law, N.~W.
\newblock 2011.
\newblock Collaborative argumentation and justifications: A statistical
  discourse analysis of online discussions.
\newblock {\em Computers in Human Behavior} 27(2):946--955.

\bibitem[\protect\citeauthoryear{Matusov and von
  Duyke}{2009}]{matusov2009bakhtin}
Matusov, E., and von Duyke, K.
\newblock 2009.
\newblock Bakhtin's notion of the internally persuasive discourse in education:
  Internal to what?
\newblock {\em Perspectives and limits of dialogism in Mikhail Bakhtin}  174.

\bibitem[\protect\citeauthoryear{McLaren, Scheuer, and
  Mik{\v{s}}{\'a}tko}{2010}]{mclaren2010supporting}
McLaren, B.~M.; Scheuer, O.; and Mik{\v{s}}{\'a}tko, J.
\newblock 2010.
\newblock Supporting collaborative learning and e-discussions using artificial
  intelligence techniques.
\newblock {\em International Journal of Artificial Intelligence in Education}
  20(1):1--46.

\bibitem[\protect\citeauthoryear{Nie, Bennett, and
  Goodman}{2019}]{nie-etal-2019-dissent}
Nie, A.; Bennett, E.; and Goodman, N.
\newblock 2019.
\newblock {D}is{S}ent: Learning sentence representations from explicit
  discourse relations.
\newblock In {\em Proceedings of the 57th Annual Meeting of the Association for
  Computational Linguistics},  4497--4510.
\newblock Florence, Italy: Association for Computational Linguistics.

\bibitem[\protect\citeauthoryear{Noroozi \bgroup et al\mbox.\egroup
  }{2018}]{noroozi2018promoting}
Noroozi, O.; Kirschner, P.~A.; Biemans, H.~J.; and Mulder, M.
\newblock 2018.
\newblock Promoting argumentation competence: Extending from first-to
  second-order scaffolding through adaptive fading.
\newblock {\em Educational Psychology Review} 30(1):153--176.

\bibitem[\protect\citeauthoryear{Oraby \bgroup et al\mbox.\egroup
  }{2017}]{oraby2017may}
Oraby, S.; Gundecha, P.; Mahmud, J.; Bhuiyan, M.; and Akkiraju, R.
\newblock 2017.
\newblock How may i help you?: Modeling twitter customer serviceconversations
  using fine-grained dialogue acts.
\newblock In {\em Proceedings of the 22nd International Conference on
  Intelligent User Interfaces},  343--355.
\newblock ACM.

\bibitem[\protect\citeauthoryear{Parker}{2006}]{parker2006public}
Parker, W.~C.
\newblock 2006.
\newblock Public discourses in schools: Purposes, problems, possibilities.
\newblock {\em Educational Researcher} 35(8):11--18.

\bibitem[\protect\citeauthoryear{Pedregosa \bgroup et al\mbox.\egroup
  }{2011}]{scikit-learn}
Pedregosa, F.; Varoquaux, G.; Gramfort, A.; Michel, V.; Thirion, B.; Grisel,
  O.; Blondel, M.; Prettenhofer, P.; Weiss, R.; Dubourg, V.; Vanderplas, J.;
  Passos, A.; Cournapeau, D.; Brucher, M.; Perrot, M.; and Duchesnay, E.
\newblock 2011.
\newblock Scikit-learn: Machine learning in {P}ython.
\newblock {\em Journal of Machine Learning Research} 12:2825--2830.

\bibitem[\protect\citeauthoryear{Pennebaker, Francis, and
  Booth}{2001}]{pennebaker2001linguistic}
Pennebaker, J.~W.; Francis, M.~E.; and Booth, R.~J.
\newblock 2001.
\newblock Linguistic inquiry and word count: Liwc 2001.
\newblock {\em Mahway: Lawrence Erlbaum Associates} 71(2001):2001.

\bibitem[\protect\citeauthoryear{Pennington, Socher, and
  Manning}{2014}]{pennington2014glove}
Pennington, J.; Socher, R.; and Manning, C.
\newblock 2014.
\newblock Glove: Global vectors for word representation.
\newblock In {\em Proceedings of the 2014 conference on empirical methods in
  natural language processing (EMNLP)},  1532--1543.

\bibitem[\protect\citeauthoryear{Pinkwart \bgroup et al\mbox.\egroup
  }{2009}]{pinkwart2009evaluating}
Pinkwart, N.; Ashley, K.; Lynch, C.; and Aleven, V.
\newblock 2009.
\newblock Evaluating an intelligent tutoring system for making legal arguments
  with hypotheticals.
\newblock {\em International Journal of Artificial Intelligence in Education}
  19(4):401--424.

\bibitem[\protect\citeauthoryear{Prasad \bgroup et al\mbox.\egroup
  }{2008}]{prasad2008penn}
Prasad, R.; Dinesh, N.; Lee, A.; Miltsakaki, E.; Robaldo, L.; Joshi, A.~K.; and
  Webber, B.~L.
\newblock 2008.
\newblock The penn discourse treebank 2.0.
\newblock In {\em LREC}.

\bibitem[\protect\citeauthoryear{Ros{\'e} \bgroup et al\mbox.\egroup
  }{2008}]{rose2008analyzing}
Ros{\'e}, C.; Wang, Y.-C.; Cui, Y.; Arguello, J.; Stegmann, K.; Weinberger, A.;
  and Fischer, F.
\newblock 2008.
\newblock Analyzing collaborative learning processes automatically: Exploiting
  the advances of computational linguistics in computer-supported collaborative
  learning.
\newblock {\em International journal of computer-supported collaborative
  learning} 3(3):237--271.

\bibitem[\protect\citeauthoryear{Salton and Buckley}{1988}]{salton1988term}
Salton, G., and Buckley, C.
\newblock 1988.
\newblock Term-weighting approaches in automatic text retrieval.
\newblock {\em Information processing \& management} 24(5):513--523.

\bibitem[\protect\citeauthoryear{Schwarz \bgroup et al\mbox.\egroup
  }{2018}]{schwarz2018orchestrating}
Schwarz, B.~B.; Prusak, N.; Swidan, O.; Livny, A.; Gal, K.; and Segal, A.
\newblock 2018.
\newblock Orchestrating the emergence of conceptual learning: A case study in a
  geometry class.
\newblock {\em International Journal of Computer-Supported Collaborative
  Learning} 13:189--211.

\bibitem[\protect\citeauthoryear{Schwarz, Neuman, and
  Biezuner}{2000}]{schwarz2000two}
Schwarz, B.~B.; Neuman, Y.; and Biezuner, S.
\newblock 2000.
\newblock Two wrongs may make a right... if they argue together!
\newblock {\em Cognition and instruction} 18(4):461--494.

\bibitem[\protect\citeauthoryear{Shum and Lee}{2013}]{shum2013politeness}
Shum, W., and Lee, C.
\newblock 2013.
\newblock (im) politeness and disagreement in two hong kong internet discussion
  forums.
\newblock {\em Journal of Pragmatics} 50(1):52--83.

\bibitem[\protect\citeauthoryear{Stolcke \bgroup et al\mbox.\egroup
  }{2000}]{stolcke2000dialogue}
Stolcke, A.; Ries, K.; Coccaro, N.; Shriberg, E.; Bates, R.; Jurafsky, D.;
  Taylor, P.; Martin, R.; Ess-Dykema, C.~V.; and Meteer, M.
\newblock 2000.
\newblock Dialogue act modeling for automatic tagging and recognition of
  conversational speech.
\newblock {\em Computational linguistics} 26(3):339--373.

\bibitem[\protect\citeauthoryear{Suthers}{2003}]{suthers2003representational}
Suthers, D.~D.
\newblock 2003.
\newblock Representational guidance for collaborative inquiry.
\newblock In {\em Arguing to learn}. Springer.
\newblock  27--46.

\bibitem[\protect\citeauthoryear{Teasley \bgroup et al\mbox.\egroup
  }{2008}]{teasley2008cognitive}
Teasley, S.; Fischer, F.; Dillenbourg, P.; Kapur, M.; Chi, M.; Weinberger, A.;
  and Stegmann, K.
\newblock 2008.
\newblock Cognitive convergence in collaborative learning.

\bibitem[\protect\citeauthoryear{Trausan-Matu, Dascalu, and
  Rebedea}{2014}]{trausan2014polycafe}
Trausan-Matu, S.; Dascalu, M.; and Rebedea, T.
\newblock 2014.
\newblock Polycafe -- automatic support for the polyphonic analysis of cscl
  chats.
\newblock {\em International Journal of Computer-Supported Collaborative
  Learning} 9(2):127--156.

\bibitem[\protect\citeauthoryear{Wise and Chiu}{2011}]{wise2011analyzing}
Wise, A.~F., and Chiu, M.~M.
\newblock 2011.
\newblock Analyzing temporal patterns of knowledge construction in a role-based
  online discussion.
\newblock {\em International Journal of Computer-Supported Collaborative
  Learning} 6(3):445--470.

\bibitem[\protect\citeauthoryear{Yoon}{2011}]{yoon2011using}
Yoon, S.~A.
\newblock 2011.
\newblock Using social network graphs as visualization tools to influence peer
  selection decision-making strategies to access information about complex
  socioscientific issues.
\newblock {\em Journal of the Learning Sciences} 20(4):549--588.

\bibitem[\protect\citeauthoryear{Zhang \bgroup et al\mbox.\egroup
  }{2018}]{zhang2018conversations}
Zhang, J.; Chang, J.; Danescu-Niculescu-Mizil, C.; Dixon, L.; Hua, Y.;
  Taraborelli, D.; and Thain, N.
\newblock 2018.
\newblock Conversations gone awry: Detecting early signs of conversational
  failure.
\newblock In {\em Proceedings of the 56th Annual Meeting of the Association for
  Computational Linguistics (Volume 1: Long Papers)},  1350--1361.

\bibitem[\protect\citeauthoryear{Zhang, Culbertson, and
  Paritosh}{2017}]{zhang2017characterizing}
Zhang, A.; Culbertson, B.; and Paritosh, P.
\newblock 2017.
\newblock Characterizing online discussion using coarse discourse sequences.

\end{thebibliography}
\bibliographystyle{aaai}
\end{document}